\documentclass[conference] {IEEEtran}
\usepackage{floatrow}
\DeclareFloatFont{normalsize}{\normalsize}
\usepackage{graphicx}
\usepackage{amsmath}
\usepackage{caption}
\title{\huge Towards Improved Human Action Recognition Using Convolutional Neural Networks and Multimodal Fusion of Depth and Inertial Sensor Data}
\author{Zeeshan Ahmad \\
	Department of Electrical and Computer Engineering\\
	Ryerson University\\
	Toronto, Canada\\
	z1ahmad@ryerson.ca
	\and
	  Naimul Mefraz Khan\\
	  Department of Electrical and Computer Engineering\\
	  Ryerson University\\
	  Toronto, Canada\\
  n77khan@ee.ryerson.ca
}

\begin{document}
	\maketitle
	\begin{abstract}
		
		This paper attempts at improving the accuracy of Human Action Recognition (HAR) by fusion of depth and inertial sensor data. Firstly, we transform the depth data into Sequential Front view Images(SFI) and fine-tune the pre-trained AlexNet on these images. Then, inertial data is converted into Signal Images (SI) and another convolutional neural network (CNN) is trained on these images. Finally, learned features are extracted from both CNN, fused together to make a shared feature layer, and these features are fed to the classifier. We experiment with two classifiers, namely Support Vector Machines (SVM) and softmax classifier and compare their performances. The recognition accuracies of each modality, depth data alone and sensor data alone are also calculated  and compared with fusion based accuracies to highlight the fact that fusion of modalities yields better results than individual modalities. Experimental results on UTD-MHAD and Kinect 2D datasets show that proposed method achieves state of the art results when compared to other recently proposed visual-inertial action recognition methods.
				
	\end{abstract} 
	\section{Introduction}
	Human action recognition is a challenging and an active research area in computer vision and machine learning, with applications in numerous fields of daily life including sports~\cite{zhou2016never}, healthcare~\cite{corbishley2008breathing}, visual surveillance~\cite{qin2016compressive}, human gesture recognition~\cite{kim2016hand}, military and robotics. \let\thefootnote\relax\footnote{© 2020 IEEE. Personal use of this material is permitted. Permission from IEEE must be obtained for all other uses, in any current or future media, including reprinting/republishing this material for advertising or promotional purposes, creating new collective works, for resale or redistribution to servers or lists, or reuse of any copyrighted component of this work in other works.}
	
	Earlier methods for HAR were statistical methods where hand crafted
	features were used for action recognition. Commonly used statistical methods involve the calculation of statistical features like mean and variance in time domain~\cite{bao2004activity} and the coefficients of  Fast Fourier transform in frequency domain~\cite{krause2003unsupervised}. 
	The first disadvantage associated with statistical method is the requirement of domain knowledge about the data~\cite{plotz2011feature}. Another disadvantage is the separation of feature extraction part from the classification part.
	Recent success of deep models in the fields of image
	classification~\cite{krizhevsky2012imagenet}, speech recognition~\cite{hinton2012deep} and their capacity to learn complex features directly from the raw data has shifted the focus of research towards deep learning.
	
	The advancement in technology, especially the release of  Microsoft Kinect, smartphones, and the availability of fast processors have brought new dynamics in HAR. HAR dataset and methods can be divided into vision based and Inertial Measurement Unit (IMU) based~\cite{cook2013transfer}. Both approaches have advantages and limitations. Although depth maps and images of vision-based approach provide reasonable accuracy for HAR, occlusion, view point variation, illumination, scaling and noise reduce the performance of this approach. IMU-based HAR is achieved by attaching a few inertial sensors to different parts of an individual's body. Although not limited by vision-based HAR problems, IMU-based approach becomes cumbersome (increase in number of sensors) and inaccurate when the underlying actions are complex. It has been recently shown that fusing the two modalities and and employing statistical machine learning techniques results in alleviating these shortcomings, which, in turns, provide improved performance in action recognition~\cite{chen2015improving}.

In this paper, we further improve visual-inertial HAR through employing novel feature transformation and a deep learning architecture. We convert depth data into Sequential Front view Images (SFI) and inertial data into signal images and through use of AlexNet and two classifiers namely SVM and softmax classifier, we achieve state of art results in terms of recognition accuracy. The key contributions of the presented work are:
\begin{itemize}
	
	\item Motivated from the fact that Convolutional Neural Networks (CNNs) are typically designed for image classification task, we convert both datasets, depth and inertial datasets into sequential front view and signal images, respectively. Conversion to image data enables abstraction through CNN to different feature types (e.g. edges, curves, and higher-level abstractions)  that are not possible with 1D temporal data~\cite{hatami2018classification}. It also enables creating a generic architecture that will be applicable to both image and non-image data without major changes.   
	
	\item Combination of deep models with statistical models  have been successful in many applications of machine learning and pattern recognition. Inspired from the results in ~\cite{xue2016cnn},~\cite{shima2017pattern} and~\cite{agarap2018neural}, we combined Convolutional Neural Network with statistical models (SVM and Softmax Classifier). We trained deep models for learning features from datasets and used statistical models for recognition task. To our knowledge, this is the first work that provides comprehensive evaluation of a deep learning-based method on benchmark datasets for visual-inertial HAR.
	
\end{itemize}
The recognition accuracies of individual modalities, depth and inertial data, are also calculated and compared against the multimodal accuracies. Experimental results on two datasets, namely UTD-MHAD and Kinect 2D prove the significance of the proposed method. We achieve average accuracies of 98.7\% and 99.8\% on these datasets, respectively, beating the state-of-the-art in visual-inertial HAR.


	\section{Related Work}
	
	Human Action Recognition from depth data and inertial sensor data have been studied actively since the advancement in technology has lead to the availability of inexpensive depth and inertial sensors. Recent studies are focused on fusing the different modalities at early level, feature level or decision level for better recognition results. Example modalities that are typically combined together are depth and inertial sensor data, RGB data and depth data, motion history images(MHI) and depth motion maps(DMM). 
	
	Comparing to other modalities, fusion of depth and inertial sensor data gained significant attention due to the cost effectiveness and availability of these sensors~\cite{chen2017survey}.
	Authors in~\cite{liu2014fusion} merge inertial and depth data to train hidden Morkov model for improving accuracy and robustness of hand gesture recognition. A real-time fusion system for human action recognition is developed in~\cite{chen2016real} by decision level fusion of depth data and inertial sensor data. An accurate and robust upper limb tracking system is developed in~\cite{tian2015upper} by unscented Kalman filter based fusion of inertial and depth sensor data. Fusion approaches show the noticeable improvement in recognition accuracy. A refined embedded system for fall detection based on KNN classifier is designed in~\cite{kwolek2015improving} by integration of depth and accelerometer data. In~\cite{hwang2017multi} features extracted by two convolutional neural networks from image and sensor data are fused together to train a Recurrent Neural Network (RNN) with Long Short Term Memory (LSTM) units for classification purpose.

	Other related work where inertial sensor data and depth data are used individually for action recognition includes the work in~\cite{gammulle2017two} where the features extracted from final convolutional layer and the first fully connected layer of CNN  are combined to train LSTM network for human action recognition.
	In~\cite{hammerla2016deep} the performance of state-of-the-art deep learning models, convolutional neural network and recurrent neural network, for human activity recognition using inertial sensors are rigorously explored by performing thousands of recognition experiments with randomly sampled model configurations to investigate the suitability of each model for HAR. Authors in~\cite{jiang2015human} converted the time series data obtained from inertial sensor into raw stacked activity images which serve as an input to CNN for human action recognition. In~\cite{ha2015multi} CNN with 2D kernel and CNN with 1D kernel are used for human activity recognition on inertial sensor dataset and their performance is compared on the basis of size of kernel.
	
	 In~\cite{chen2013real} depth video sequence is projected onto three orthogonal Cartesian planes to generate depth motion maps corresponding to front, top and side views and then use these DMMs as features to train collaborative representation classifier. The work in~\cite{yang2012recognizing} involves the generation of Histograms of Oriented Gradients (HOG) from depth motion maps to build DMM-HOG descriptors for human action recognition. Comprehensive analysis of multimodal fusion of user-generated multimedia content is explained in~\cite{shah2017multimodal}.
	 
	 In previous vision-inertial fusion methods, features are extracted directly from raw data without transformation. In proposed method we transform depth and inertial data into sequential front view and signal images respectively. Transforming data to images empower CNN to pull out the different features like edges, corner and patches that would be impossible with raw data.

	Our work is inspired from~\cite{chen2015improving}~\cite{hwang2017multi},~\cite{gammulle2017two}, and~\cite{jiang2015human} as explained in the next section.
	
	\begin{figure*}[h]
		\centering
		\includegraphics[width=\linewidth]{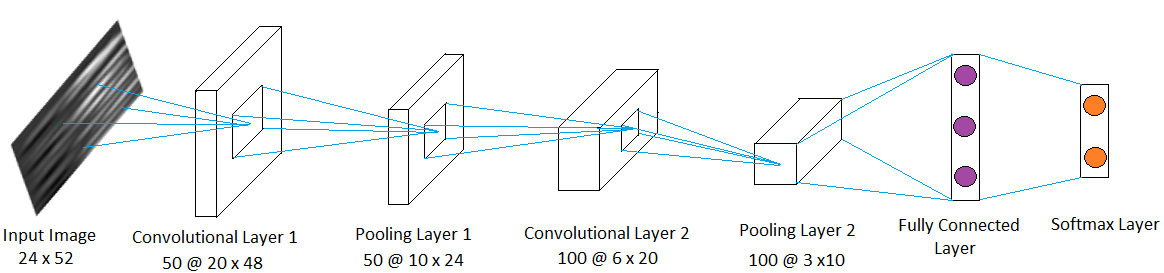}
		\caption{CNN Architecture for Signal Image}
		\label{fig:CNN Architecture}
		
	\end{figure*}

	\begin{figure*}
		\vspace{2cm}
		\centering
		\includegraphics[width=\linewidth]{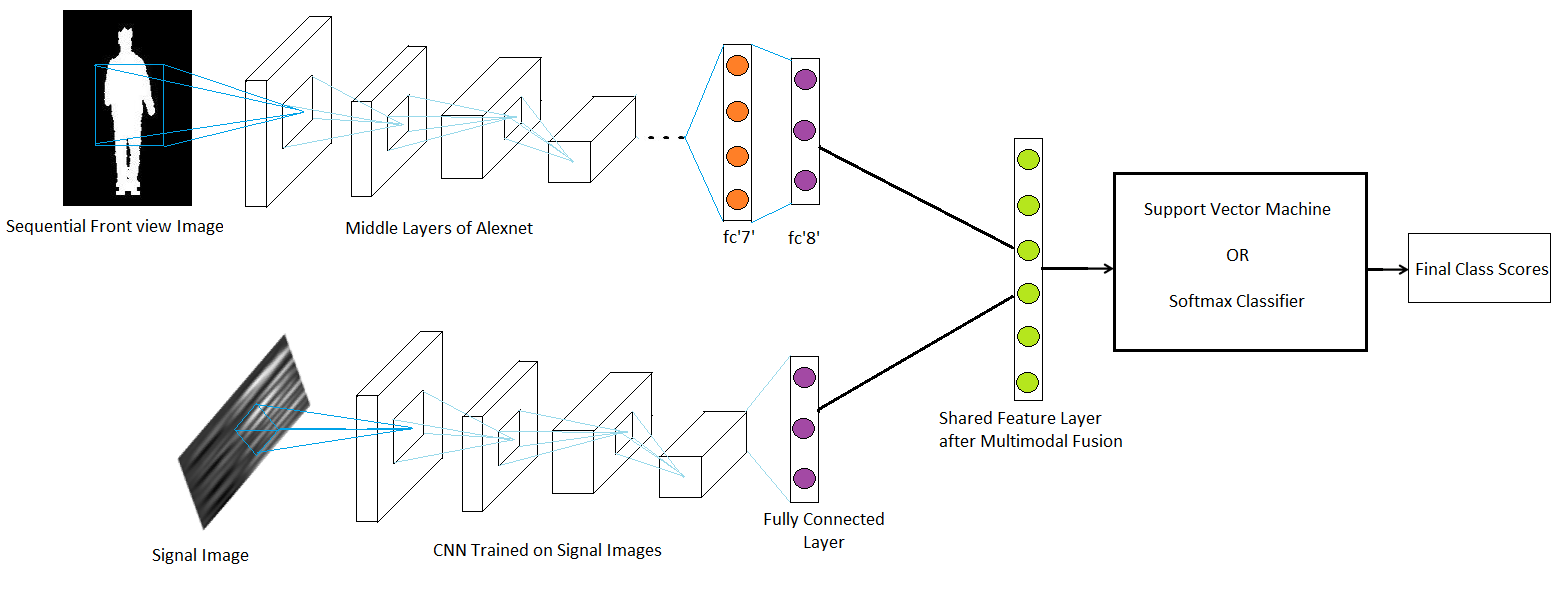}
		
		\caption{Complete Overview of Proposed Method using Multimodal Fusion of Depth and Inertial Data}
		\label{fig: overview}
	\end{figure*}
	\begin{figure*}[h]
		\centering
		\includegraphics[width=\linewidth]{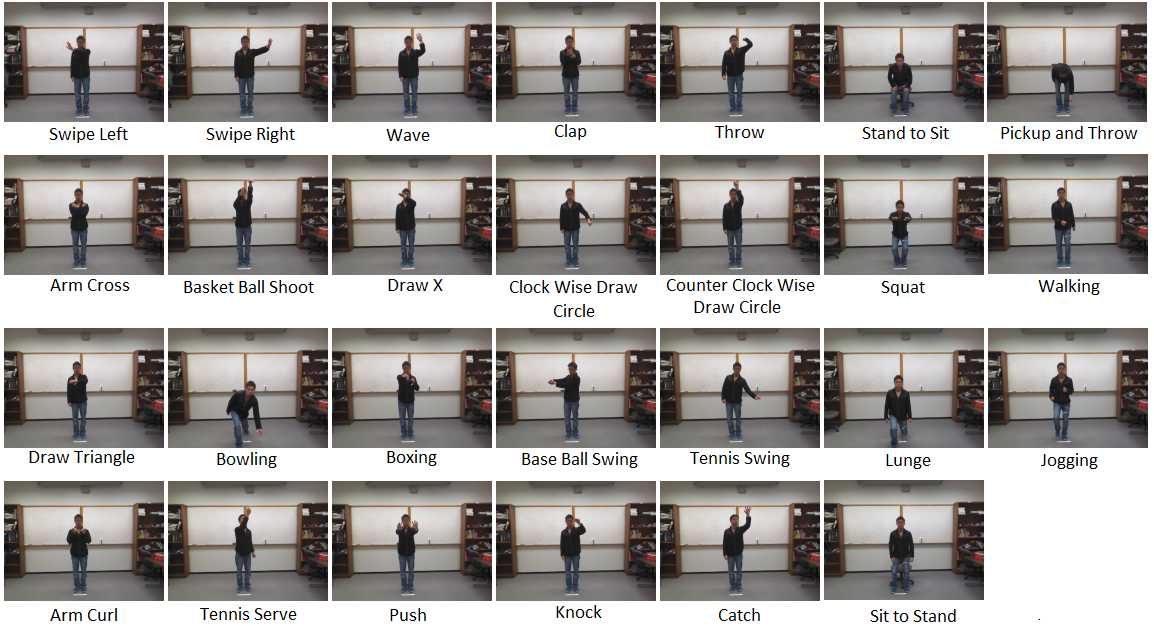}
		\caption{Sample actions from the UTD-MHAD data set}
		\label{fig:Samples}
	\end{figure*}

	\section{Proposed Method} \label{proposed method}
	
	In our proposed method we convert the depth data into Sequential Front view Images(SFI) and inertial data into signal images. We fine-tune the pre-trained AlexNet(CNN based model)~\cite{krizhevsky2012imagenet} on SFIs to leverage the full potential of transfer learning and train another CNN, whose architecture is shown in Figure~\ref{fig:CNN Architecture}, on signal images. Finally, learned features are extracted from the fully connected layers of both CNNs, fused to make a shared layer of features, and fed as training / testing data to a supervised classifier. We experiment with two classifiers : support vector machines (SVM) and softmax classifier. The overview of proposed method is shown in Figure~\ref{fig: overview}. The benefits of transfer learning and formation of SFIs and signal images are explained in detail below.

	\subsection{Transfer Learning}
	The concept of transfer learning is derived from multi-task learning~\cite{caruana1998multitask} where the purpose is to transfer the already acquired knowledge on a particular domain to different but related domain.
	Transfer learning is very important in those computer vision, data mining and machine learning applications where the datasets are small but could be made compatible for transfer learning models. AlexNet is a CNN model that has been popular for transfer learning purpose. Transfer learning on pre-trained AlexNet is executed by fine tuning the model on new dataset and doing some fine tuning involving the adjustment of some layers especially to adjust the last fully connected layers according to the number of classification categories of the input dataset. 
		\begin{figure}[h]
		\centering
		\includegraphics[width=\linewidth]{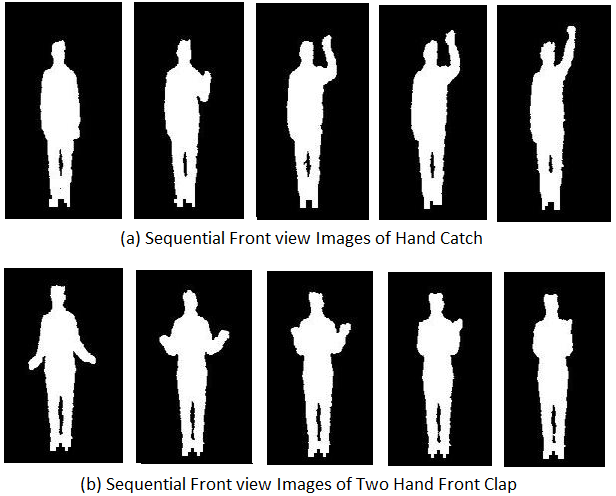}
		\caption{Sequential Front View Images}
		\label{fig:first-figure}
	\end{figure}
	
	\subsection{Classifiers}
	In the proposed method the classifiers used are Support Vector Machines and softmax classifier. These classifiers are explained in detail below.
	\subsubsection{Support Vector Machine}
	Support Vector Machines (SVM) are famous supervised classifiers for data classification, face recognition, feature and text categorization and regression~\cite{burges1998tutorial}.
	In simplest form, the score function for SVM is the mapping of the input vector to the scores and is a simple matrix operation as shown in Equation~\ref{eq:svm}.
	\begin{equation}\label{eq:svm}
	f=Wx + b
	\end{equation}
	Where $x$ is the input vector, $W$ is the weight determined by input vector and the number of classes and $b$ is the bias vector.
	When used for multimodal fusion, the input to the classifier is the concatenated  feature vector from individual modalities or the scores generated by each classifier.
	In~\cite{wang2009multimodal} and~\cite{bouzouina2017multimodal}, score level fusion based on SVM is performed for human authentication by combining the  features obtained from face and iris verifiers.
	\subsubsection{Softmax Classifier}
	Softmax classifier is a multiclass classifier or regressor used in the fields of machine learning, data mining, mathematics, statistics and allied fields~\cite{wolfe2017application}.
	Score function for softmax classifier computes the class specific probabilities whose sum is 1.
	
	The mathematical representation of score function for softmax classifer is shown below.
	\begin{equation}\label{eq:soft}
	f(y)=\frac{e^{y_j}}{\sum_ke^{y_k}}
	\end{equation}
	where $y$ is the input vector and the score function maps the exponent domain to the probabilities. 
	
	When applied to  multimodal fusion, the input to the classifier is from shared layer which is formed by combining the features from various modalities.
	
	\subsection{Formation of Sequential Front View Images}
	In~\cite{yang2012recognizing}, the front, side and top view of depth motion maps are generated from depth sequences. From these views we observe that using all three views of DMMs are not necessary, since we have supplemental information from inertial dataset. Thus we convert the depth sequences into front view images called Sequential Front view Images (SFI) as shown in Figure~\ref{fig:first-figure}. By using SFIs, we are reducing the computational cost of the experiment. These images are similar to the motion energy images and motion history images introduced in~\cite{bobick2001recognition}.  These SFIs provide cumulative information about the action from start to completion.
	 
	\subsubsection{Preprocessing}
	The SFIs are converted to 3-channel images and resized to 227x227 using bicubic interpolation to be compatible with AlexNet.
	\subsubsection{Fine Tuning}
	To apply Alexnet to our classification problem, few modifications are required. We reduce the size of fully connected layer 'fc8' from 1000 to size equal to the number of classes in our datasets and then replace the output classification layer with new classification layer suited for our datasets.
	\begin{figure}[h]
		\centering
		\includegraphics[width=\linewidth]{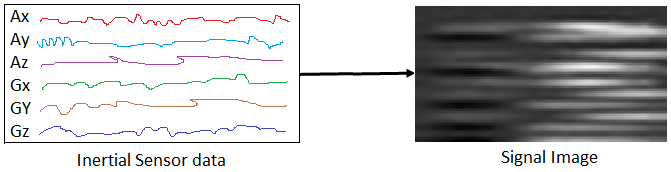}
		\caption{Conversion of Inertial Sensor Data into Signal Image}
		\label{fig:conversion}
	\end{figure}

	\subsection{Formation of Signal Images}
	The data obtained from inertial sensors is in the form of multivariate time series. The wearable devices in our datasets are tri-axis accelerometers and gyroscopes. Hence we have six sequences of signals : three accelerometer and three angular velocity sequences. These sequences are converted into 2D virtual images called signal images as shown in Figure~\ref{fig:conversion}. With the given six signal sequences, signal image is obtained through row-by-row stacking of raw signal sequences in such a way that each sequence comes adjacent to every other sequence based on the algorithm in~\cite{jiang2015human}. The signal images are formed by taking advantage of the temporal correlation among the signals. The row-by-row stacking of six sequences has the following order.
	
\textit{1234561352461425361526161}
	
	 Where the numbers \textit{1} to \textit{6} represents the sequence numbers in a raw signal. From the above order of the sequences, it is observed that every sequence neighbors every other sequence to make a signal image. Thus we have 25 rows of sequences in our signal image. By removing the last row of the signal image, the final width of our signal image becomes \textbf{24}. 
	 
\begin{figure}[h]
		\centering
		\includegraphics[width=\linewidth]{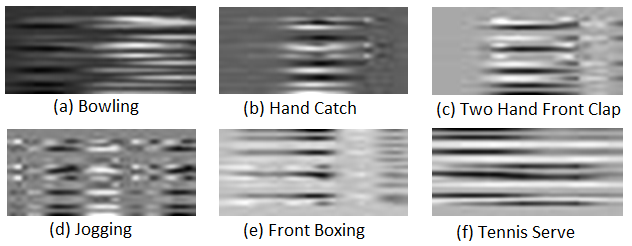}
		\caption{Signal Images of size 24x52}
		\label{fig:signal images}
	\end{figure}

	 Majority of human actions falls in low frequency band, therefore the length of the signal image  can be selected equal to sampling rate of dataset which is 50Hz for our datasets. Since the shortest sequence in our datasets has only 107 samples, therefore in order to incorporate all the samples into a signal image and to facilitate the design of CNN of Figure~\ref{fig:CNN Architecture}, the length of the signal image is finalized as \textbf{52}. Thus the final size of our signal image is \textbf{24x52} as shown in Figure~\ref{fig:signal images}.

	\subsubsection{CNN Architecture for Signal Images}
	The architecture of CNN proposed for performing action recognition task on signal images consists of two convolutional layers, two pooling layers, a fully connected layer and a softmax layer as shown in Figure~\ref{fig:CNN Architecture}. The first convolutional layer has 50 kernels of size 5x5, followed by pooling layer of size 2x2 and stride 2. The output of the first pooling layer is the input of the second convolutional layer which has 100 kernels followed by 2x2 pooling layer with stride 2. The last two layers are fully connected layers and softmax layer.
	
	\subsection{Fusion}
	 In the last part of proposed method, we extract the learned features from the last fully connected layers of both CNNs, integrate the features to form a single feature layer which serves as input to our classification network to perform the action recognition task as shown in Figure~\ref{fig: overview}.

	\section{Experiments and Results}
	We experiment on publicly available UTD Multimodal Human Action Dataset (UTD-MHAD)~\cite{chen2015utd} and Kinect 2D dataset~\cite{blog} by separately training and testing three different input data types : SFIs alone, signal images alone and their combination. We used subject specific setting for experiments on both datasets as the subject specific setting generates better recognition results due to less intraclass variations~\cite{chen2016fusion}. We conduct our experiments on Matlab R2018a on a desktop computer with NVIDIA GTX-1070 GPU.
	
	\subsection{UTD-MHAD Dataset}\label{MHAD dataset}
	The UTD-MHAD dataset contains both depth and inertial data components and  consists of 27 samples of different actions as shown in Figure~\ref{fig:Samples}.
	The first part of experiment is to fine tune AlexNet on the SFIs obtained from the depth sequences of dataset. The number of SFIs obtained from depth data are enough for transfer learning. We select 46636 samples for retraining AlexNet model and 11660 samples for testing. We retrain AlexNet for 50 epochs. We perform experiment 20 times by randomly selecting the same percentage of training and testing samples : 46636 samples for training and 11660 samples for testing and report the average accuracy. The values of other training parameters are shown below in Table~\ref{tab:parameters for alexnet}.
	\begin{table}[h]
		\centering
		\begin{tabular}{|c|c|}
			
			\hline 
			\textbf{Training Parameters} & \textbf{Values}   \\\hline 
			Momentum  &      0.9 \\\hline
			Initial Learn Rate  &      0.005 \\\hline
			Learn Rate Drop Factor  &      0.5 \\\hline
			Learn Rate Drop Period  &      10 \\\hline
			$L_2$ Regularization  &      0.004 \\\hline
			Max Epochs  &      50\\\hline
			MiniBatchSize  &   128 \\\hline		
			
		\end{tabular}
		\caption{Training Parameters for AlexNet}
		\label{tab:parameters for alexnet}
	\end{table}
	We reached these values through using the grid search method. For example, we evaluate AlexNet for values [0.01, 0.05, 0.001,
	0.005] of initial learning rate. Similarly Momentum for the values of [ 0.7, 0.8 , 0.9] and so on.

	The second part of the experiments is to train the CNN on signal images obtained from inertial data. The inertial sensor component of UTD-MHAD dataset is very challenging to train a CNN. The first deficiency is that inertial sensor was worn either on volunteer's right wrist or right thigh depending upon nature of action. Hence the  sensor is worn only on two positions for collecting data of 27 actions which is not enough to capture all the dependencies and characteristics of data. The other challenge is that the number of data samples is very small to train and test the CNN as compared to the inertial dataset in~\cite{shoaib2014fusion} where sensors were worn at five different places and there are large number of samples. Thus the number of signal images obtained from inertial sensor data are only 1722.
	
	To overcome these problems we perform data augmentation on signal images to increase the number of samples.

	\subsubsection{Data Augmentation}\label{data augmentation}
	Signal Images are 2D virtual images formed by row by row stacking of signals by preserving the correlation among the rows of the signal. Hence only those data augmentation techniques are valid that could not alter the correlation among the rows of the signal image. Following data augmentation techniques are used:
	\paragraph{Flipping}
	we flip signal images left to right and upside down : reverse the original image across horizontal axis and vertical axis. Flipping of image preserve the correlation and doesn't alter the size of the signal image.
\begin{figure*}[t]
	\centering
	\includegraphics[width=\linewidth]{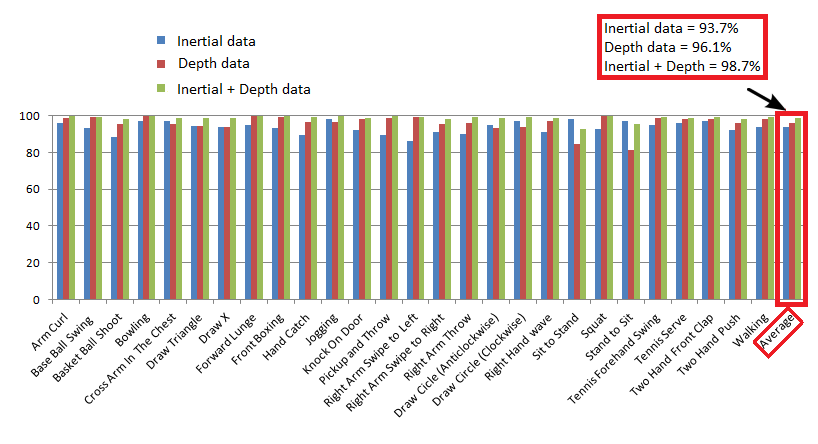}
	\caption{Class Specific accuracies of UTD-MHAD data set for Depth data only, Inertial Data only and Fusion of Depth and Inertial Data}
	\label{fig:Class accuracies}
\end{figure*}

	\paragraph{Rotation}
	Rotating image at a particular angle is an important data augmentation technique. A negative angle represents a clockwise rotation and a positive angle represents a counter clockwise rotation. Care should be taken while selecting the angle so that the dependencies among the signal samples do not change. We rotate the signal images to 180 degrees to get more samples of the images.
	\paragraph{Adding Gaussian Noise}
	To further increase the number of training samples we add Gaussian Noise of zero mean and variance of 0.009.
	
	After augmentation we get 13776 samples of signal images. We select 11021 for training and 2755 for testing. We train CNN on signal images as shown in Figure~\ref{fig:CNN Architecture} for 100 epochs. We perform experiment 20 times by randomly selecting the same percentage of training and testing samples : 11021 samples for training and 2755 samples for testing and report the average accuracy. The values of remaining parameters for training CNN are shown below in Table~\ref{tab:parameters for CNN}.
	\begin{table}[h]
		\centering
		\begin{tabular}{|c|c|}
			
			\hline 
			\textbf{Training Parameters} & \textbf{Values}   \\\hline 
			Momentum  &      0.9 \\\hline
			Initial Learn Rate  &      0.001 \\\hline
			Learn Rate Drop Factor  &      0.5 \\\hline
			Learn Rate Drop Period  &      10 \\\hline
			$L_2$ Regularization  &      0.004 \\\hline
			Max Epochs  &      100\\\hline
			MiniBatchSize  &   64 \\\hline		
			
		\end{tabular}
		\caption{Training Parameters for CNN}
		\label{tab:parameters for CNN}
	\end{table}

	In the final part of the experiment, learned features are extracted from the fully connected layers of both CNNs, combined together and input to the  classification network. We work with two classification networks :  multiclass SVM and softmax classifier. We achieve the average accuracies of 98.7\% and 98.2\% respectively with the classification networks.
	
	The results obtained by proposed method and their comparison to previous methods for UTD-MHAD dataset are shown in Table~\ref{tab:comparisonTabe}. As we can see, the proposed fusion-based method with SVM beats the current state-of-the-art~\cite{mahjoub2018efficient} by 0.2\%, which proves
	the value of the proposed method. Although the improvement is marginal, at such high level of accuracy even a small boost is significant. We also show the class specific accuracies in Figure~\ref{fig:Class accuracies} when the modalities are used independently and fused. As expected, fusion results in better accuracy for all classes. Comparing SVM with softmax, we see that SVM performs better.  Softmax classifier reduces the cross-entropy function  while SVM employs a margin based function. Multiclass SVM classifies data by locating the hyperplane at a position where all data points are classified correctly. Thus SVM determines the maximum margin among the data points of various classes~\cite{tang2013deep}. The more rigorous nature of classification is likely the reason why SVM performs better than softmax.

	\begin{table}[h]
		\centering
		\begin{tabular}{|c|c|}
			
			\hline 
			\textbf{Previous Methods} & \textbf{Accuracy\%}  \\\hline 
			C.chen et al.~\cite{chen2015utd}       &      79.1 \\\hline
			Bulbul et al.~\cite{bulbul2015dmms}    &      88.4 \\\hline
			J.Imran et al.~\cite{imran2016human}   &      91.2 \\\hline
			Chen et al.~\cite{chen2016real}        &      97.2 \\\hline
			Mahjoub et al.~\cite{mahjoub2018efficient}        &      98.5 \\\hline
				\multicolumn{2}{|l|}{\textbf{\textit{Proposed Method}}}\\
				\hline
			Depth + Inertial Sensor (Softmax Classifier) & 98.2 \\\hline
			\textbf{Depth + Inertial Sensor (SVM Classifier)}  & \textbf{98.7} \\
			\hline				
		\end{tabular}
		\caption{ Comparison of Accuracies of proposed method with previous methods on UTD-MHAD dataset}
		\label{tab:comparisonTabe}
	\end{table}
	
	\subsection{kinect2D Dataset}
	
	Kinect 2D action dataset is another publicly available dataset that contains both depth and inertial data. It is a new dataset using the second generation of Kinect~\cite{chen2016fusion}. It contains 10 actions performed by six subjects with each subject repeating the action 5 times. The 10 actions are "right hand high wave", "right hand catch", "right hand high throw", "right hand draw X", "right hand draw tick", "right hand draw circle", "right hand horizontal wave", "right hand forward punch", "right hand hammer, and "hand clap".
	
	The proposed method is  applied on this dataset with the same  manner as that of UTD-MHAD dataset. The training parameters for training AlexNet on SFIs and CNN on signal images are same as that of Tables~\ref{tab:parameters for alexnet} and ~\ref{tab:parameters for CNN}, however the number of  training and testing samples are different from UTD-MHAD dataset. The total SFIs obtained from the depth data sequences are 17622. We select 14098 samples for training AlexNet and 3524 for testing.
	
\begin{figure}[h]
	\centering
	\includegraphics[width=\linewidth]{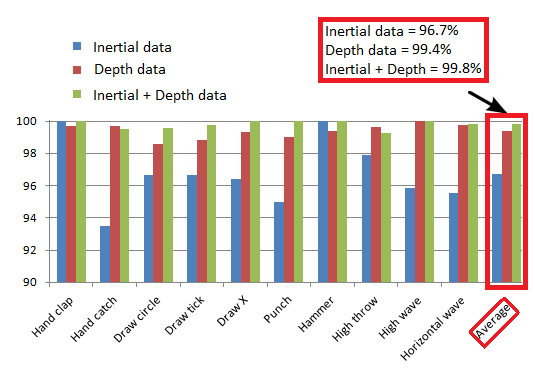}
	\caption{Class Specific accuracies of Kinect 2D dataset}
	\label{fig:Class accuracies of 2D dataset}
\end{figure}

	The signal images obtained from inertial data are fewer in number to train the CNN of Figure~\ref{fig:CNN Architecture}. Thus the same data augmentation techniques explained in section ~\ref{MHAD dataset} are used on signal images for increasing the number of samples. After data augmentation we select 3533 samples for training and 833 samples for testing. Experimental results on this dataset and their comparison with previous method are shown below in Table~\ref{tab:results on 2D dataset}. The class specific accuracies are shown in Figure~\ref{fig:Class accuracies of 2D dataset}.
	
	\begin{table}[h]
		\centering
		\begin{tabular}{|c|c|}
			
			\hline 
			\textbf{Previous Methods} & \textbf{Accuracy\%}  \\\hline 
			Chen et al.~\cite{chen2016fusion}       &      99.5 \\\hline
			\multicolumn{2}{|l|}{\textbf{\textit{Proposed Method}}}\\
			\hline
			Depth + Inertial (softmax classifier)  &     99.5 \\\hline
			\textbf{	Depth + Inertial (SVM classifier)}  &  \textbf{99.8} \\\hline			
		\end{tabular}
		\caption{Comparison of Accuracies of proposed method with previous methods on Kinect 2D dataset}
		\label{tab:results on 2D dataset}
	\end{table}

Table~\ref{tab:results on 2D dataset} shows the  accuracy of 99.8\% with SVM classifier beating the previous state of art 99.5\%~\cite{chen2016fusion}. We see that even for Kinect 2D dataset, SVM performs better than softmax. 

We achieved higher accuracies with Kinect 2D dataset as compare to UTD-MHAD dataset. It is due to the fact that interclass discrimination in Kinect 2D dataset is higher than UTD-MHAD dataset and none of the action has class specific accuracy less than 93\% as shown in figure~\ref{fig:Class accuracies of 2D dataset}. On the other hand, there are actions in UTD-MHAD dataset which are less discriminant such as "sit to stand" and "stand to sit" and "right arm swipe to left" and "right arm swipe to right". The class specific accuracies of these less discriminant actions are low as compare to other actions as shown in figure~\ref{fig:Class accuracies}.
	\section{Conclusion}
	In this paper, we perform a novel multimodal fusion scheme for visual-inertial HAR using CNN, SVM and softmax classifier. We successfully extracted learned features from fully connected layers of CNNs and perform the recognition task by SVM  or softmax classifier. Our experiments on UTD-MHAD and Kinect 2D datasets achieves state of the art results. We outperform the current state-of-the-art on UTD-MHAD dataset by 0.2\% , with an increase from 98.5\% to 98.7\% and on Kinect 2D dataset by 0.3\%, with an increase from 99.5\% to 99.8\%. In this paper we did not utilize skeleton data due to jitters in skeleton joint positions in UTD-MHAD dataset captured by the first generation of Kinect. In our future work, we are planning to apply the proposed method on more datasets, combine other modalities than depth and inertial sensors, explore other fusion methods and employ end-to-end deep learning architectures. 
	\bibliographystyle{IEEEtran}

\end{document}